\documentclass[runningheads]{article}

\usepackage{graphicx}
\usepackage{todonotes}
\usepackage{listings}
\usepackage{amsmath}
\usepackage{url}
\usepackage{enumitem}

\usepackage{framed}

\usepackage[utf8]{inputenc}   
\usepackage{graphicx}

\title{Kaapana: A Comprehensive Open-Source Platform for Integrating AI in Medical Imaging Research Environments}

\author{%
Ünal Akünal\textsuperscript{1}\thanks{Contributed equally. Each co-first author may list themselves as lead author on their CV.}%
\and
Markus Bujotzek\textsuperscript{1,2}\footnotemark[1]%
\and
Stefan Denner\textsuperscript{1,3}\footnotemark[1]%
\and
Benjamin Hamm\textsuperscript{1,2}\footnotemark[1]%
\and
Klaus Kades\textsuperscript{1}\footnotemark[1]%
\and
Philipp Schader\textsuperscript{1,3}\footnotemark[1]%
\and
Jonas Scherer\textsuperscript{1}\footnotemark[1]%
\and
Marco Nolden\textsuperscript{1,4}%
\and
Peter Neher\textsuperscript{1,4,5}%
\and
Ralf Floca\textsuperscript{1,6}%
\and
Klaus Maier-Hein\textsuperscript{1,4,5,7}%
}

\date{} 

\begin{document}
\maketitle

\begin{center}
\textsuperscript{1} Division of Medical Image Computing, German Cancer Research Center, Heidelberg, Germany\\
\textsuperscript{2} Medical Faculty, University of Heidelberg, Heidelberg, Germany\\
\textsuperscript{3} Computer Science Faculty, University of Heidelberg, Heidelberg, Germany\\
\textsuperscript{4} Pattern Analysis and Learning Group, Department of Radiation Oncology, Heidelberg University Hospital, Heidelberg, Germany\\
\textsuperscript{5} German Cancer Consortium (DKTK), Partner Site Heidelberg, Heidelberg, Germany\\
\textsuperscript{6} Heidelberg Institute of Radiation Oncology (HIRO), National Center for Radiation Research in Oncology (NCRO), Heidelberg, Germany\\
\textsuperscript{7} National Center for Tumor Diseases (NCT), NCT Heidelberg, a partnership between DKFZ and the University Medical Center Heidelberg, 69120 Heidelberg, Germany
\end{center}

\paragraph{Motivation}
Recent advancements in AI-driven medical imaging research have enabled automated analysis of complex datasets, offering significant potential for clinical insights and scientific discovery \cite{esteva2021medicalcv}.  However, there remains a substantial gap between methodological innovation and its real-world implementation, largely due to the vast amounts of data required to develop generalizable and robust AI models \cite{sokol2025translation}. This gap arises primarily from fundamental barriers such as limited access to medical data and strict regulatory constraints, which restrict data sharing and centralization.
A paradigm to overcome these challenges is \textit{bringing algorithms to data} rather than pooling data centrally \cite{guan2024fedsurvey,pati2022flbigdata,willemink2020preparing}.  
Research software plays a critical role in this approach by providing standardized and scalable infrastructure suited to clinical environments. 
Existing solutions often fall short of meeting practical needs \cite{diaz2021data}. Although some platforms offer integrated solutions with unified interfaces, these are frequently narrowly tailored to specific use cases. Their typically monolithic architectures and often restrictive licensing models limit adaptability, impeding the reuse of components across diverse projects or domains.
Consequently, research teams typically resort to fragmented toolchains, resulting in significant technical overhead due to a lack of integration, scalability, and user-friendliness \cite{becker2024streamline}.
This fragmented landscape not only slows down the research process but also raises the risk of errors and reduces reproducibility. Each stage of the medical imaging AI pipeline — data ingestion, preprocessing, model training, and post-processing — typically relies on separate tools, making the workflow inefficient and siloed. Moreover, the lack of a unified, intuitive user interface further separates clinical and technical collaborators, stifling interdisciplinary synergy that is essential for translational research.

These issues become even more pronounced in multi-center studies, which are vital for developing robust AI models capable of generalizing across diverse patient populations and clinical practices \cite{pati2022flbigdata}. Multi-center studies inherently account for variations in medical procedures, imaging protocols, and patient demographics, thereby significantly enhancing the clinical utility and universal applicability of AI solutions. Additionally, many diseases fall into the long-tail category, characterized by low incidence rates at individual institutions, making single-center studies insufficient. Multi-center collaborations are thus crucial for aggregating sufficient data to reliably train and validate models for these rare conditions. However, conducting multi-center research introduces further complexities such as increased data heterogeneity, logistical challenges, and heightened privacy and regulatory considerations \cite{willemink2020preparing}.

To truly harness the promise of AI in medical imaging, there is a pressing need for a flexible, modular, and user-centric platform that bridges the gap between methodological research and clinical applicability. Such a platform must support secure, scalable, and collaborative workflows across institutional boundaries and empower both data scientists and clinicians to contribute effectively to the full lifecycle of medical AI research.

\paragraph{Kaapana is the Solution.}

To overcome the challenges of fragmented workflows, limited data access, and regulatory constraints, we present Kaapana: a versatile, open-source platform designed to unify and streamline medical imaging research. Kaapana integrates the entire imaging AI pipeline into a single, modular infrastructure, enabling researchers and clinicians to collaborate more effectively across institutional boundaries.

Kaapana provides a standardized, interoperable environment that integrates seamlessly with diverse clinical IT systems. This \textit{bring the algorithm to the data} paradigm does not only support scalable AI deployment, but also addresses regulatory compliance by minimizing data movement and ensuring patient privacy.

With built-in federated processing capabilities, Kaapana also facilitates secure, distributed training and validation of machine learning models on sensitive, institution-specific datasets. This allows researchers to harness the full diversity of clinical data while preserving privacy — enabling robust, generalizable AI development and supporting large-scale, multi-institutional collaboration.

The platform's modular architecture, built on containerized microservices, ensures high adaptability to different research needs. Each component can be independently configured, replaced, or extended, supporting a wide range of use cases without vendor lock-in. By building on proven, industry-standard open-source technologies, Kaapana offers both security and scalability by design. By enabling open access to the source code, Kaapana also encourages the developer community to strengthen, customize and enhance the tools and solutions provided inside the platform.

Crucially, Kaapana provides an end-to-end pipeline for data ingestion, curation, processing, and result interpretation — all accessible through a unified interface. This integrated workflow reduces fragmentation, lowers technical overhead, and fosters interdisciplinary collaboration between data scientists and clinicians, accelerating innovation in medical imaging AI.

\begin{figure}[ht!]
 \centering
 \includegraphics[width=\textwidth]{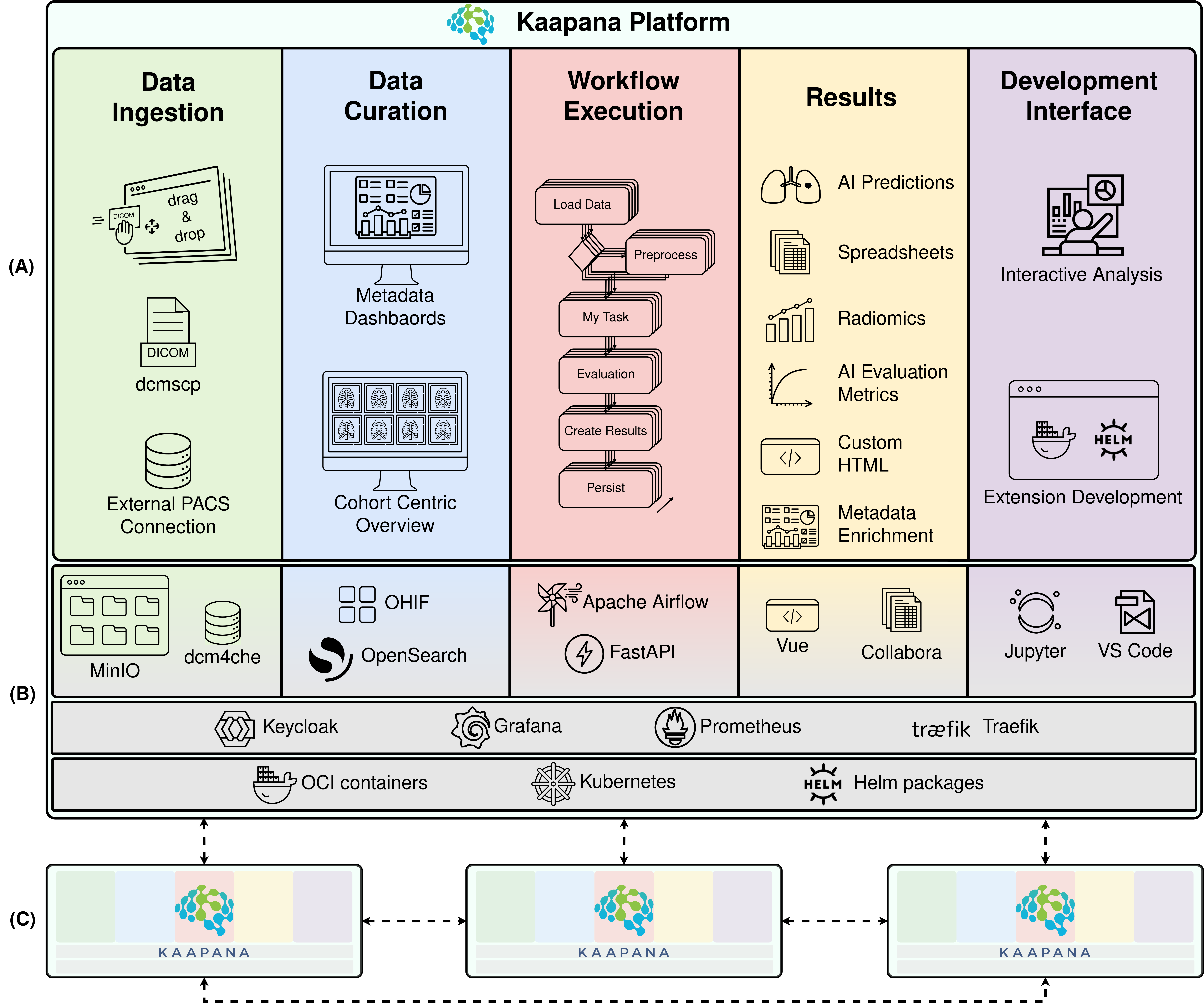}
 \caption{To overcome challenges in multi-institutional AI research, including fragmented toolchains, regulatory constraints, and lack of standardized infrastructure; Kaapana offers a modular and open-source solution. (A) The platform unifies the medical image analysis workflow: Data Ingestion, Curation, Workflow Execution, Results Analysis, and Development Interface, thereby enabling clinical and data science collaboration. (B) This integration is enabled by a robust, open-source tech stack supporting interoperability, scalability, and reproducibility across environments. (C) Kaapana instances can operate collaboratively to support multi-center studies and federated learning, allowing AI models to be trained across institutions while preserving data sovereignty.}
 \label{fig1:kaapana}
\end{figure}

\paragraph{Technical Overview.}

Kaapana is built on a robust microservice architecture, with Kubernetes at its core to orchestrate containerized services and workflows. This design enables scalable, flexible, and consistent infrastructure across diverse research environments ranging from single institutions to multi-site federations.

Deployment is facilitated through Helm charts, which encapsulate all necessary configurations for both on-premise and cloud-based clusters. This approach ensures portability and simplifies integration into existing clinical IT systems, regardless of underlying infrastructure.

To support reproducibility and ease of maintenance, Kaapana includes a dedicated build system that automates the generation of deployable artifacts such as Helm charts, container images, and custom scripts. The system resolves dependencies and performs automated sanity checks, ensuring reliable installations across various environments. Deployments can be performed directly from an OCI-compliant registry in connected environments or through pre-packaged builds in offline settings. This enables the platform to adapt to institutional network policies and operational constraints.

Security and access control are integral to Kaapana’s design. The platform employs Keycloak and an OAuth2 Proxy for fine-grained authentication and authorization, with seamless integration into site-local LDAP directories. This ensures secure, role-based access while aligning with institutional data governance and compliance requirements.

Kaapana delivers complete end-to-end workflow coverage within a single, unified platform. From data ingestion and curation to processing, visualization, and analysis, all functionality is accessible through an intuitive graphical user interface. This eliminates the need for switching between disjointed tools, reducing operational overhead and fostering seamless collaboration between clinical and technical stakeholders.

\paragraph{Kaapana's End-to-End Processing Pipeline.}

Kaapana provides a comprehensive, modular pipeline within a secure and unified environment that supports the full spectrum of medical imaging research from data ingestion to clinical result interpretation.

The platform supports robust ingestion of imaging data in both DICOM and NIfTI formats. Data can be imported via the DICOM DIMSE protocol or through a user-friendly drag-and-drop web interface into Kaapana’s integrated PACS. It also enables seamless communication with clinical PACS systems for both sending and receiving DICOM images, requiring minimal configuration. For non-imaging data, Kaapana uses MinIO’s S3-compatible object storage to ensure flexibility and scalability.

Data exploration and cohort curation are facilitated through intuitive interfaces. The Datasets View module offers interactive thumbnail browsing for rapid screening and selection of radiological images, while DICOM metadata are automatically indexed using OpenSearch and made accessible via a metadata dashboard. This dashboard supports exploration of attribute distributions and detection of potential cohort biases. Advanced filtering and tagging capabilities further streamline cohort refinement for downstream processing \cite{denner2023efficient}.

At the heart of the processing pipeline is a workflow management system powered by Apache Airflow, which models imaging workflows as directed acyclic graphs (DAGs). Each DAG node represents an operator that can range from simple Python scripts to containerized services. This abstraction allows the system to remain agnostic to specific languages or tools while still accommodating diverse processing tasks, from data pre-processing to advanced AI-driven analysis such as segmentation, radiomics, and classification.

Kaapana also supports federated and distributed computing that allows collaboration between sites without transferring raw data. Multiple instances can be linked to enable secure cross-site workflow execution and federated model training. This architecture maintains institutional data sovereignty while harnessing collective computational and research resources \cite{Kades2022}.

Operator reusability is another key feature of the platform. Prebuilt modules for common tasks such as data loading, conversion, and output storage enable rapid workflow assembly without redundant development. State-of-the-art AI pipelines, including TotalSegmentator \cite{Wasserthal_2023} and nnU-Net \cite{Isensee2021}, are integrated out of the box, minimizing setup time and enabling rapid deployment of advanced analyses.

A dedicated view for workflow results consolidates workflow outputs, including visual and quantitative results. Image-based outputs, such as predicted segmentations, are stored in the PACS (Picture archiving and communication system) and can be reviewed via the Datasets View. Tabular and statistical results are exportable to analytical environments like JupyterLab and Collabora, supporting in-depth evaluation and reporting.

Overall, Kaapana’s end-to-end pipeline consolidates data ingestion, curation, analysis, and result visualization into a single cohesive platform that streamlines radiological workflows and enables scalable, collaborative medical imaging research.

\paragraph{Extensibility and Customizability.}

Kaapana is designed with extensibility and customizability at its core, empowering users to adapt the platform to a wide range of research scenarios. Users can readily incorporate existing processing operators or develop new ones tailored to their specific workflows. To support this, Kaapana includes an Extension Development Kit (EDK) that simplifies the development and deployment of new components. This is especially valuable in restricted environments where external data access is limited or compute resources are constrained.

Extensions are packaged as Helm charts, allowing for flexible distribution and deployment. In connected (online) environments, extensions can be pulled from an OCI-compliant registry, while in offline settings, they can be uploaded directly through a drag-and-drop interface. This dual-mode distribution model ensures that the platform remains operable and extensible across diverse infrastructure conditions.

Beyond processing workflows, Kaapana supports seamless integration of custom web-based services, which can be connected to core platform components. This adaptability is exemplified by Kaapana’s native support for specialized image viewers: OHIF Viewer for radiological images and SLIM Viewer for pathology data. Both viewers interface directly with the PACS system for visualization. Furthermore, the platform supports streaming of desktop applications within the web interface via pre-configured noVNC containers. Tools such as MITK and 3D Slicer can thus be integrated effortlessly, enabling users to continue working with familiar, domain-specific applications.

In addition to these extension capabilities, Kaapana’s modular architecture allows users to tailor the entire platform to their specific needs. Unnecessary components can be excluded to reduce resource usage, enabling lightweight deployments in resource-constrained environments. Thanks to the layered design of its components, users can customize individual modules with minimal disruption to overall functionality or interoperability. This enables both expansion and reduction of the platform, ensuring Kaapana can scale effectively with the requirements of varying use cases, compute environments, and user preferences.

\paragraph{Real-World Applications and Impact.}

Kaapana has been successfully adopted across a range of national and international research initiatives, demonstrating its flexibility, scalability, and real-world impact. Notable deployments include the RACOON network\footnote{\url{https://racoon.network}}, the CCE-DART consortium\footnote{\url{https://cce-dart.com}}, the German Cancer Consortium (DKTK) \cite{Scherer2020}\footnote{\url{https://dktk.dkfz.de}}, and the NeuroRad project\footnote{\url{https://stroke.ccibonn.ai}} \cite{brugnara2023deep}. These examples span diverse clinical and research contexts, validating Kaapana's utility across different domains.

Kaapana’s federated processing capabilities were pivotal in large-scale, privacy-preserving collaborations. In RACOON and CCE-DART, distributed workflows enabled secure analysis of medical imaging data across multiple institutions without centralizing sensitive data \cite{bujotzek2025real}. Within the RACOON-COMBINE project, Kaapana’s capacity for national-scale deployment was highlighted by facilitating the extraction of clinically relevant findings from a federated patient cohort encompassing all German university hospitals.

Kaapana’s customizable features were further showcased in the NeuroRad project \cite{brugnara2023deep}, where the platform was streamlined and extended with a domain-specific interface tailored to stroke imaging. This use case underscores Kaapana’s ability to adapt to specialized clinical workflows.

In addition to these deployments, Kaapana’s open-source nature has cultivated an active and growing community of users and contributors. This community-driven development model not only accelerates innovation but also ensures ongoing improvement \cite{cardoso2022monai} and sustainability which broadens the platform’s impact and encourages adoption across diverse medical research environments.

\paragraph{Call to Action.}

The future of medical imaging research hinges on platforms that are not only technically robust and interoperable but also inherently privacy-preserving. Kaapana rises to this challenge by unifying AI-driven imaging workflows within an open-source, end-to-end research environment. With its modular design, federated processing capabilities, and seamless integration into clinical IT systems, Kaapana effectively bridges the gap between cutting-edge AI research and real-world clinical application.

We invite the broader research and clinical imaging community to explore, adopt, and contribute to Kaapana. By collaboratively evolving the platform, we can accelerate innovation, foster interoperability, and address the dynamic needs of medical image computing. For latest information, maintained source code, and documentation, please visit the Kaapana repository\footnote{\url{https://github.com/kaapana/kaapana}} and documentation\footnote{\url{https://kaapana.readthedocs.io}}.

\paragraph{Acknowledgments}

\begingroup
\sloppy
The Kaapana project was supported by multiple research initiatives and funding bodies. The Building Data Rich Clinical Trials (CCE\_DART) project received funding from the European Union’s Horizon~2020 research and innovation programme under grant agreement No.~965397. Additional support was provided by the Deutsche Forschungsgemeinschaft (DFG, German Research Foundation) through project No.~410981386, which focuses on capturing tumor heterogeneity in hepatocellular carcinoma.
\par
\endgroup
The work further benefited from partial support through the Data Science Driven Surgical Oncology Project (DSdSO), funded by the Surgical Oncology Program at the National Center for Tumor Diseases (NCT), Heidelberg, a partnership of DKFZ, UKHD, and Heidelberg University. Additional resources were made available by the Joint Imaging Platform of the German Cancer Consortium.

We also acknowledge partial support from the HiGHmed Consortium, funded by the German Federal Ministry of Education and Research (BMBF, funding code 01ZZ1802A), and from the RACOON project within the Netzwerk Universitätsmedizin (NUM), funded by BMBF (funding code 01KX2021). Finally, this work was supported in part by the Helmholtz Association through the Trustworthy Federated Data Analytics (TFDA) project (funding number ZT-I-OO1~4).

We are grateful to Rajesh Baidya, Mikulas Bankovic, Jens Beyermann, Lorenz Feineis, Maximilian Fischer, Hanno Gao, Lisa Kausch, Jasmin Metzger, Kaushal Parekh, Santhosh Parampottupadam, and Jonas Reinwald for their inspiration and contributions to Kaapana.

\bibliographystyle{unsrt}
\bibliography{bib}

\end{document}